\documentclass[lettersize,journal]{IEEEtran}
\usepackage{amsmath,amsfonts}
\usepackage{algorithmic}
\usepackage{algorithm}
\usepackage{array}
\usepackage[caption=false,font=normalsize,labelfont=sf,textfont=sf]{subfig}
\usepackage{textcomp}
\usepackage{stfloats}
\usepackage{url}
\usepackage{verbatim}
\usepackage{graphicx}
\usepackage{cite}
\usepackage{amssymb}
\usepackage{xcolor}
\usepackage{orcidlink}

\usepackage{tabularx}
\usepackage{booktabs}
\usepackage{multirow}
\usepackage{makecell}
\definecolor{deepgreen}{RGB}{0, 120, 0}  
\definecolor{deepred}{RGB}{198, 14, 14}
\usepackage{adjustbox}

\begin{document}

\title{LE4Mob: Towards Inductive, Distance-Aware and General-Purpose Location Embedding for Human Mobility Modelling}

\author{
Xinglei Wang\,\orcidlink{0000-0002-9824-7663},
Stephen Law\,\orcidlink{0000-0003-3184-572X},
Zichao Zeng\,\orcidlink{0009-0002-8975-875X},
Junyuan Liu\,\orcidlink{0009-0009-7194-6868},
Guangsheng Dong\,\orcidlink{0000-0001-7676-497X},
and Tao Cheng\,\orcidlink{0000-0002-5503-9813}%
\thanks{This work was supported by the project ``Understanding the Impact of Covid-19 \& NPIs on Mobility and Places in Singapore,'' funded by The Alan Turing Institute and DSO National Laboratories, Singapore.}
\thanks{Xinglei Wang, Zichao Zeng, Junyuan Liu, and Tao Cheng are with SpaceTimeLab, Department of Civil, Environmental and Geomatic Engineering, University College London, London, United Kingdom. Corresponding author: Tao Cheng (e-mail: tao.cheng@ucl.ac.uk).}
\thanks{Stephen Law is with the Department of Geography, University College London, London, United Kingdom, and the Department of Geography and Resource Management, The Chinese University of Hong Kong, Hong Kong, China.}
\thanks{Zichao Zeng is also affiliated with 3DIMPACT, Department of Civil, Environmental and Geomatic Engineering, University College London, London, United Kingdom.}
\thanks{Guangsheng Dong is with the State Key Laboratory of Information Engineering in Surveying, Mapping and Remote Sensing (LIESMARS), Wuhan University, Wuhan, China. He is also a visiting scholar with SpaceTimeLab, Department of Civil, Environmental and Geomatic Engineering, University College London, London, United Kingdom.}

\thanks{Tao Cheng is also with The Alan Turing Institute, London, United Kingdom.}
}


\maketitle

\begin{abstract}
Location representations provide mobility models with fundamental information about the spatial position, functional characteristics, and relationships of places. However, existing embeddings are often dependent on mobility observations, unable to represent unseen locations, and weakly constrained to retain geographic distance. This limits their reuse across datasets and mobility tasks. To address these limitations, we propose LE4Mob, an inductive, distance-aware, and geography-derived location embedding framework for mobility modelling. LE4Mob extends contrastive language–location pre-training 
while introducing a distance-aware regularisation objective that encourages the embedding space to preserve spatial relationships. Pre-trained from geographic context, LE4Mob can encode rich spatial-semantic information and generate embeddings for unseen locations inductively. Its independence from downstream mobility task supervision also makes it transferable across different mobility tasks. We evaluate LE4Mob on individual-level next location prediction and population-level commuter flow generation. Experiments across multiple datasets and study areas show that LE4Mob outperforms strong baselines, with particular advantages in inductive settings and when downstream models rely directly on interactions between location embeddings. These findings demonstrate the potential of distance-aware, geography-derived location representations as reusable foundations for human mobility modelling. 


\end{abstract}


\section{Introduction}
Human mobility, the movement of individuals across space and time, shapes urban dynamics, transport demand, disease transmission, economic activity, and public policy \cite{barbosa2018human}. The increasing availability of large-scale mobility data, together with advances in deep learning, has led to rapid progress in computational mobility modelling and established data-driven methods as an important direction for mobility science \cite{pappalardo2023future}. Within this broad field, four major tasks have received particular attention: next location prediction, trajectory generation, crowd flow prediction, and flow generation \cite{luca2021survey}. These tasks differ in their prediction targets, spatial scales, and methodological formulations, but they share a foundational requirement: models must represent the locations between, through, or towards which movement occurs.

Locations are commonly encoded as dense vectors that provide compact inputs to downstream models. Effective representations should describe not only location identity, but also geographic position, surrounding urban functions, and spatial relationships with other places. These properties are central to mobility behaviour. Individuals choose destinations partly according to the activities and opportunities available there \cite{stouffer1940intervening,schneider1959gravity}, while travel likelihood and aggregate interaction volumes are strongly constrained by geographic distance \cite{zipf1946p}. Location representation can therefore be considered as a common foundation for both individual- and population-level mobility modelling.

Nevertheless, location representations have largely been developed separately for different task families. Individual-level models commonly learn embeddings from historical check-ins or trajectories, using co-occurrence or transition patterns to infer relationships among destinations \cite{feng2017poi2vec,zhao2017geo,zhou2018deepmove,wan2021pre,lin2021pre}. Population-level models often represent regions using handcrafted attributes, such as population, land use, and points of interest (POIs) \cite{simini2021deep}, or learn latent representations from observed origin–destination (OD) networks\cite{liu2020gmel,yin2023convgcnrf}. These approaches can capture task-relevant patterns, but the resulting embeddings are usually tied to the mobility data, spatial units, and downstream objective used during training.

This task-specific paradigm creates three limitations. First, mobility-derived embeddings may inadequately capture the functional semantics of the urban environment. Human movement is influenced by the spatial distribution and accessibility of activity opportunities (e.g., employment, education, recreation, transport services) associated with different land uses and urban functions \cite{cervero1997travel,lee2015relating}. Although semantic attributes can be introduced as auxiliary features, they are not necessarily encoded in a reusable representation when locations are learned mainly from their occurrence in mobility sequences or OD graphs.

Second, many existing embeddings are transductive. They assign a trainable vector to each location observed during training and cannot directly encode unseen destinations or regions. This constraint is problematic when new locations, spatial units, or study areas appear after deployment. Extending the location vocabulary then requires retraining or learning additional embeddings, limiting the transfer across datasets and regions.

Third, existing location representations rarely preserve geographic distance explicitly. 
Distance is a fundamental constraint on both destination choice and aggregate flows, as reflected in distance-decay effects and classical gravity-\cite{zipf1946p} and radiation-based formulations \cite{simini2012universal}. However, embeddings optimised for mobility co-occurrence, semantic similarity, or graph connectivity do not necessarily retain geographic proximity. Functionally similar but distant locations may be placed close together, whereas nearby locations with different functions may become separated. Such distortion can make it harder for downstream models to capture interaction patterns. Some recent work has recognised the importance of geographic distance and introduced soft distance supervision \cite{han2026spatially}, but it focuses on street-view image representations for geolocalisation rather than location embeddings for mobility modelling.

These limitations motivate distance-aware geography-driven location representation learning. Instead of deriving embeddings from a particular mobility dataset, a location encoder can learn from information available independently of downstream mobility observations, such as coordinates, POIs, text, imagery. Such an encoder should be distance-explicit, inductive, and generalisable across mobility tasks. 



Our previous work provided initial evidence for this direction by applying CaLLiPer \cite{wang2025multi}, a POI-based spatial-semantic location encoder, to next location prediction \cite{wang2025into}. CaLLiPer uses contrastive language–location pre-training to align coordinate representations with textual descriptions of surrounding POIs. Its embeddings improved next location prediction, particularly when individuals visited locations unseen during downstream training. However, this study considered only individual-level task, and CaLLiPer did not explicitly preserve geographic distance in its embedding space.

To address these limitations, we propose LE4Mob, an inductive, semantics-rich, and distance-aware \textbf{\underline{L}}ocation \textbf{E}mbedding framework \textbf{for} human \textbf{Mob}ility modelling. LE4Mob extends CaLLiPer with a distance-preserving objective that aligns relationships in the embedding space with geographic distances between locations. Its pre-training relies on general geographic information rather than trajectory or OD-flow labels, allowing the same encoder to be integrated into different mobility datasets and model architectures.

We evaluate LE4Mob on two different mobility modelling tasks: individual-level next location prediction and population-level commuter flow generation. These tasks differ in spatial units, prediction targets, and modelling assumptions, providing a stringent test of the model’s generalisability across heterogeneous mobility settings. Across conventional and inductive next location prediction, LE4Mob achieves the strongest overall performance, with particularly clear advantages in the inductive setting and on geographic distance-based errors. In commuter flow generation in London, LE4Mob achieves 2.2--11.3\% performance improvements relative to the strongest baselines. Ablation and qualitative analyses further demonstrate that the distance-preserving design improves the spatial awareness and downstream utility of the learned location embeddings.


The contributions are threefold. Conceptually, we frame location representation learning as a transferable problem spanning individual- and population-level mobility modelling. Methodologically, we propose LE4Mob, a framework that addresses task dependence, transductivity, and spatial distortion by combining semantic context, inductive encoding, and distance-aware regularisation. Empirically, we evaluate LE4Mob on next location prediction and commuter flow generation, demonstrating its effectiveness across distinct mobility scales, and release the code and datasets to support reproducibility.


\section{Related work}

In this study, \textbf{location representation} is an umbrella term for any numerical description of a geographic location, including raw attributes, handcrafted features, and learned latent representations. A \textbf{location embedding} refers specifically to a dense latent vector, either assigned to a fixed location identifier through a lookup table or generated from observable geographic attributes by a location encoder. \textbf{Location representation learning} denotes the broader process of learning these embeddings or the encoder that produces them. 

The geographic entities regarded as locations vary by task. Individual-level models commonly represent locations as discretised spatial cells, POIs, clustered stay points or significant places \cite{hong2023context}. Population-level models generally use areal units such as grids, traffic analysis zones, census tracts, output areas \cite{simini2021deep}, etc.  Despite the differences, the locations in both models can be characterised by their geographic position, functional or semantic attributes, and spatial relationships with other locations. 

Accordingly, a general-purpose location representation should satisfy four properties. It should capture semantic and functional characteristics because travel decisions depend on the activities and opportunities available at a location. It should preserve spatial relationships that shape destination choice and aggregate interactions. It should also generalise inductively to locations unseen during representation learning and remain transferable across downstream objectives. The following subsections examine how existing approaches address these requirements in location encoding, individual mobility prediction, and population flow modelling.

\subsection{Location Encoders}

Location encoders are parameterised functions that map observable geographic information to dense representations. Coordinate-based encoders typically transform longitude and latitude through positional encoding and a neural network \cite{mai2022review}. Because they learn a mapping rather than a location-specific lookup table, they can encode arbitrary coordinates and are inherently inductive.

Some encoders are trained through supervised classification \cite{mai2020iclr,mai2023sphere2vec}, whereas multimodal methods use contrastive learning to align coordinates with images, text, POIs, or other environmental observations \cite{klemmer2025satclip,wang2025multi,liu2025enriching}. The latter capture both geographic position and urban function and may therefore transfer beyond their original training tasks.

CaLLiPer \cite{wang2025multi} is a multimodal geography-derived encoder that aligns coordinate representations with textual descriptions of surrounding POIs. Its objective is independent of trajectories and OD flows, enabling application across mobility datasets. However, incorporating coordinates does not guarantee that the final embedding geometry preserves geographic distance. Nonlinear transformations optimised for classification or contrastive alignment may place semantically similar but distant locations close together or separate nearby places with different functions. Few geographic encoders explicitly supervise the correspondence between geographic and embedding space distances, although this property is particularly important for mobility modelling.

\subsection{Location Representations for Individual Mobility Prediction}

Next location prediction estimates a person’s next destination from previous movements. Deep models commonly represent candidate destinations using trainable lookup embeddings and combine them with temporal, user, and contextual features \cite{feng2018deepmove,hong2023context}. These embeddings are efficient but depend entirely on task-specific observations, do not inherently encode coordinates or urban functions, and cannot represent locations outside the training vocabulary.

Self-supervised approaches derive richer embeddings from movement sequences. Word2Vec-style methods treat locations as words and trajectories as sentences \cite{mikolov2013efficient}, while extensions incorporate spatial \cite{feng2017poi2vec}, temporal \cite{wan2021pre}, and spatial-temporal information \cite{zhao2017geo}. BERT-inspired models further produce context-dependent representations from adjacent visits \cite{lin2021pre}. These approaches capture behavioural regularities, but sparse or unseen locations receive unreliable or no representations. Moreover, spatial co-occurrence does not necessarily reflect functional semantics or geographic proximity of locations.


A previous work instead encodes coordinates and environmental context independently of trajectories, supporting inductive prediction for unseen destinations \cite{wang2025into}. However, it did not explicitly preserve geographic distance or examine population-level transfer.

\subsection{Location Representations for Population Flow Modelling}

Population flow modelling estimates aggregate movements between regions, including commuting, migration, and travel demand \cite{barbosa2018human}. Flow prediction forecasts future flows from historical observations, whereas flow generation estimates an origin--destination (OD) matrix from the attributes and relationships of origins and destinations \cite{simini2021deep}. This study focuses on the latter.

Classical gravity models relate flows to the masses of origin and destination regions and their spatial separation \cite{zipf1946p}, while radiation models emphasise the role of intervening opportunities \cite{stouffer1940intervening,simini2012universal}. DeepGravity replaces the rigid functions with neural networks to capture nonlinear interactions \cite{simini2021deep}. Recent models further embed spatial interaction principles into deep architectures \cite{zhao2026theory}.
These methods primarily concern downstream flow modelling rather than the pre-training of transferable representations of the regions.

A growing body of work instead learns latent location representations jointly with a downstream flow decoder, such as a multilayer perceptron \cite{tu2026gagnn} or bilinear interaction model \cite{liu2020gmel,xu2025predicting}. Many such approaches use graph neural networks to integrate regional attributes with geographic, transport, or other relational structures, while optimising the resulting embeddings using observed OD flows as supervision \cite{liu2020gmel,yin2023convgcnrf,shi2024fusion,tu2026gagnn}. However, they are commonly evaluated under a transductive protocol in which all regions are present in a fixed graph and the split is performed over OD edges or flow records. Test region embeddings are therefore generated within the same graph used during training, and the evaluation measures prediction of unobserved flows between known regions rather than generalisation to previously unseen spatial units. Consequently, these studies provide limited evidence that the learned representations can be applied to regions absent during representation learning.

Geography-derived self-supervised methods provide an alternative by learning region representations from information available independently of OD flows, such as coordinates, POIs, land use, text, or satellite imagery. For example, recent work pre-trains region representations from satellite imagery and evaluates them using held-out origin regions \cite{xu2025predicting}. Because such representations can be generated without observing the target region’s flow records, they are better suited to encoding unseen regions and to settings where historical mobility observations are unavailable or incomplete.

Nevertheless, geographic distance is often introduced only at the downstream stage as an explicit OD-pair feature. Although effective, supplying distance to a decoder does not ensure that the region embeddings themselves retain spatial relationships. Representation-level distance supervision instead incorporates spatial structure during pre-training, producing embeddings that remain spatially informative when used with decoders that do not receive distance separately and potentially improving transfer across downstream architectures.

\section{Notation and Problem Formulation}
\label{sec:problem_formulation}

This section defines the common notation, formulates the two downstream mobility tasks, and specifies the location representation learning problem addressed by LE4Mob.

\subsection{Locations}

\textbf{Definition 1 (Location).}
Let $\mathcal{L} = \{l_i\}_{i=1}^{N}$ denote a set of $N$ locations. Each location $l_i$ is characterised by
$l_i = (g_i, c_i)$, where $g_i$ denotes its geometry and $c_i$ denotes its contextual information, such as surrounding points of interest (POIs) or land use characteristics. We further use $\mathbf{x}_i$ to denote a representative coordinate derived from $g_i$, such as the coordinate of a point location or the centroid of a polygon.

The spatial form of a location depends on the mobility task. In next location prediction, a location may correspond to a POI, venue, clustered stay region, or discretised spatial cell. In flow generation, locations are areal units forming a tessellation of the study area, such as regular grids, census zones, or administrative regions.

The geographic distance between locations $l_i$ and $l_j$ is denoted by $d_{ij} = d_{\mathrm{geo}}(\mathbf{x}_i, \mathbf{x}_j)$, where $d_{\mathrm{geo}}(\cdot,\cdot)$ is the selected geographic distance function.

\subsection{Next Location Prediction}

\textbf{Definition 2 (Individual Mobility Trajectory).}
A user’s mobility history is represented as a spatio-temporal trajectory $S_u$. It consists of a time-ordered sequence of visits:
$S_u^{1:T_u}={[(l_{u}^k,t_{u}^k)]}_{k=1}^{T_u}$
where $(l_{u}^k,t_{u}^k)$ means that user $u$ visited location $l_{u}^k \in \mathcal{L}$ at time step $t_{u}^k$.

\medskip
\textbf{Problem 1 (Next Location Prediction).}
Given a user’s previous visits within an observation window $m:n$, $S_u^{m:n}=\left[(l_u^m,t_u^m),\ldots,(l_u^n,t_u^n)\right]$, the goal is to predict the next location $l_u^{n+1}$ that the user will visit.
 The task is formulated as a multi-class classification problem. The model assigns a probability to each candidate location in the candidate location set $\mathcal{L}^{c}$, and the location with the highest probability is selected as the predicted next location: \begin{equation} \hat{l}_u^{n+1} = \arg\max_{l\in\mathcal{L}^{c}}
\mathrm{P}\left(l \mid S_u^{m:n}\right).
\end{equation}
In the inductive setting, the candidate location set $\mathcal{L}^{c}$ may include locations that were not observed during model training.

\subsection{Flow Generation}

\noindent\textbf{Problem 2 (Flow Generation).}
Let $\mathcal{L}$ denote a set of spatial regions, and let $y_{ij} \geq 0$ denote the observed flow from origin $l_i$ to destination $l_j$, where $i \neq j$. The total outflow from origin $l_i$ is $O_i = \sum_{j \neq i} y_{ij}$. Given $O_i$ and information describing the origin and destination locations, the objective is to estimate the flow $\hat{y}_{ij}$ for arbitrary location pairs $(l_i,l_j)$.

A flow generation model may estimate either $\hat{y}_{ij}$ directly or a destination probability $\hat{p}_{ij}$, from which the flow is obtained as    $\hat{y}_{ij} = O_i \hat{p}_{ij}, \sum_{j \neq i} \hat{p}_{ij} = 1$.

\subsection{Location Representation Learning for Mobility Modelling}

\noindent\textbf{Problem 3 (Location Representation Learning).}
The objective is to learn a parameterised mapping function $F_{\theta}$,
which generates a dense embedding $\mathbf{e}_i = F_{\theta}(g_i,c_i), \mathbf{e}_i \in \mathbb{R}^{d}$,
from the geographic and contextual information of location $l_i$.

The mapping function is designed to satisfy three properties:

\begin{enumerate}
    \item \textbf{Inductive.}
    The encoder $F_{\theta}$ can represent a location not observed during representation pre-training without learning a new location-specific parameter or retraining the encoder.

    \item \textbf{Distance-aware.}
    Geographic distance is explicitly incorporated into representation learning, such that relationships between embeddings retain information about the corresponding geographic distances $d_{ij}$.

    \item \textbf{Downstream-task-agnostic.}
    The parameters $\theta$ are learned independently of task-specific trajectory labels or OD flow values. The resulting embeddings can therefore be integrated into different downstream mobility models, including those defined in Problems~1 and~2.
\end{enumerate}

\section{Methodology}
\label{sec:methodology}

This section presents LE4Mob, a spatial-semantic location encoder augmented with explicit distance-aware regularisation. We first show why the spatial structure introduced by positional encoding is not necessarily retained after neural transformation. We then describe the semantic and distance-aware training objectives and the integration of the pre-trained embeddings into downstream mobility models.

\subsection{Theoretical Motivation}
\label{sec:theoretical_motivation}

Standard location encoders commonly follow
\begin{equation}
    \mathbf{e}(\mathbf{x})
    =
    f_{\theta}\bigl(\mathrm{PE}(\mathbf{x})\bigr),
\end{equation}
where $\mathbf{x}$ is a coordinate, $\mathrm{PE}(\cdot)$ is a positional encoding, and $f_{\theta}(\cdot)$ is a trainable neural network \cite{mai2022review}. Although positional encodings introduce spatial structure, an unconstrained neural transformation is not guaranteed to retain it.

\medskip
\textbf{Definition 3 (Distance Preservation).}
Let $(\mathcal{X},d_{\mathrm{geo}})$ be a geographic space and let
$\mathbf{e}:\mathcal{X}\rightarrow\mathbb{R}^{d}$ be a non-zero embedding function. We call $\mathbf{e}$ distance-preserving if, for any anchor $\mathbf{x}_i$ and locations $\mathbf{x}_j$ and $\mathbf{x}_k$,
\begin{equation}
    d_{\mathrm{geo}}(\mathbf{x}_i,\mathbf{x}_j)
    <
    d_{\mathrm{geo}}(\mathbf{x}_i,\mathbf{x}_k)
    \Longrightarrow
    \langle\mathbf{e}_i,\mathbf{e}_j\rangle
    >
    \langle \mathbf{e}_i,\mathbf{e}_k \rangle,
\end{equation}
where $\langle \cdot,\cdot \rangle$ denotes inner product or cosine similarity \cite{mai2022review}.

\medskip
\noindent\textbf{Proposition 1.}
A positional encoding may be distance-preserving, while its composition with an unconstrained linear transformation is not.

\medskip
\noindent\textit{Proof.}
Consider the one-dimensional positional encoding
\begin{equation}
    \mathrm{PE}(x)
    =
    \begin{bmatrix}
        \cos x \\
        \sin x
    \end{bmatrix},
    \qquad x\in[0,\pi].
\end{equation}
Its cosine similarity satisfies $\mathrm{PE}(x_i)^{\top}\mathrm{PE}(x_j)=\cos(x_i-x_j)$, which decreases monotonically with $|x_i-x_j|$ over this interval.

Now apply
\begin{equation}
    W=
    \begin{bmatrix}
        10 & 0 \\
        0 & 1
    \end{bmatrix}
\end{equation}
and define
\begin{equation}
    \tilde{\mathbf{e}}(x)
    =
    \frac{W\mathrm{PE}(x)}
    {\lVert W\mathrm{PE}(x)\rVert_2}.
\end{equation}
Let $x_i=\frac{\pi}{3}, x_j=\frac{\pi}{2}, x_k=0$.
Although $|x_i-x_j|=\frac{\pi}{6}<\frac{\pi}{3}=|x_i-x_k|$, the transformed similarities are $\operatorname{sim}
    \left(
        \tilde{\mathbf{e}}(x_i),
        \tilde{\mathbf{e}}(x_j)
    \right)
    =
    \frac{\sqrt{3}}{\sqrt{103}}
    \approx 0.171$, $\operatorname{sim}
    \left(
        \tilde{\mathbf{e}}(x_i),
        \tilde{\mathbf{e}}(x_k)
    \right)
    =
    \frac{10}{\sqrt{103}}
    \approx 0.985$.
    
The more distant location therefore becomes more similar to the anchor than the closer location, violating distance consistency. This failure mode is not unique to sinusoidal encodings. Even a simple distance-preserving PE like a constrained identity map is easily distorted by the anisotropy introduced by unconstrained linear layers.


In a deeper encoder trained only through semantic alignment, geographically distant but functionally similar places may be pulled together, while nearby places with different functions may be separated. To mitigate the spatial distortion, LE4Mob therefore introduces an explicit distance-aware objective.

\subsection{LE4Mob Framework}
\label{sec:framework}

\begin{figure*}[t] 
    \centering
    \includegraphics[width=1.01\textwidth]{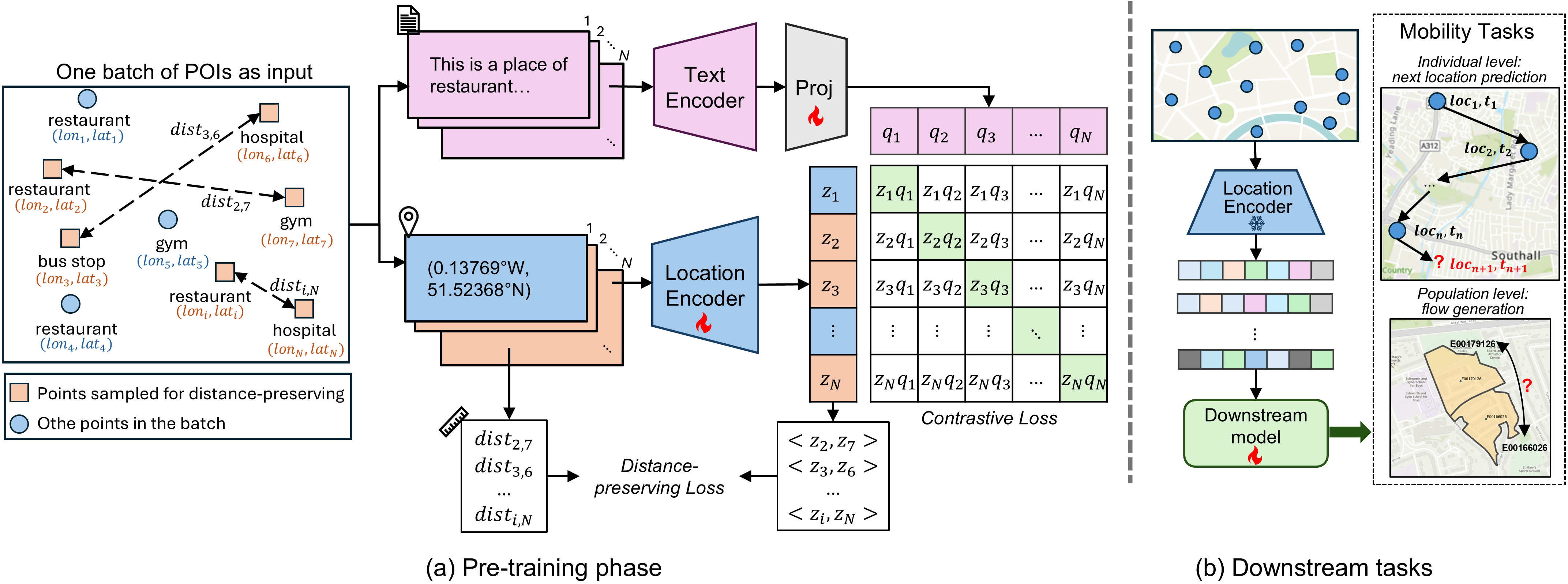}
    \caption{Framework of LE4Mob. (a) In the pre-training phase, batches of POIs are encoded through a dual-stream architecture that aligns coordinate-based location embeddings with POI-derived textual semantics using contrastive learning, while a distance-preserving loss regularises sampled location pairs to retain geographic distance information. (b) After pre-training, the frozen location encoder generates embeddings for downstream mobility tasks, including individual-level next location prediction and population-level flow generation.}
    \label{fig:framework}
\end{figure*}

LE4Mob follows a ``pre-training$\rightarrow$downstream application" paradigm, as illustrated in Fig.~\ref{fig:framework}. It extends CaLLiPer \cite{wang2025multi} and contains three backbone components:

\begin{enumerate}
    \item a location encoder $f_{\theta}$ that maps a PE-encoded coordinate $\mathbf{x}_i$ to a location embedding;
    \item a frozen text encoder $g_{\phi}$ that extracts semantic features from the contextual description $c_i$ of surrounding POIs; and
    \item a trainable projection layer $p_{\psi}$ that maps text features to the location embedding space.
\end{enumerate}

For each location $l_i$, the two branches produce
\begin{align}
    \mathbf{z}_i
    &=
    \frac{f_{\theta}(\mathrm{PE}(\mathbf{x}_i))}
    {\lVert f_{\theta}(\mathrm{PE}(\mathbf{x}_i))\rVert_2}, \\
    \mathbf{q}_i
    &=
    \frac{p_{\psi}(g_{\phi}(c_i))}
    {\lVert p_{\psi}(g_{\phi}(c_i))\rVert_2}.
\end{align}

Contextual descriptions are used only as semantic supervision during pre-training. Once training is complete, $g_{\phi}$ and $p_{\psi}$ are discarded, and $f_{\theta}$ independently generates embeddings from coordinates.

\subsection{Semantic Alignment}
\label{sec:semantic_alignment}

Given a mini-batch of $B$ paired coordinates and contextual descriptions, we calculate the cross-modal similarity
\begin{equation}
    a_{ij}
    =
    \frac{\mathbf{z}_i^{\top}\mathbf{q}_j}{\tau},
\end{equation}
where $\tau$ is a temperature parameter.

Following CLIP-style contrastive learning \cite{radford2021clip}, the location-to-text and text-to-location losses are
\begin{align}
    \mathcal{L}_{\mathrm{l2t}}
    &=
    -\frac{1}{B}
    \sum_{i=1}^{B}
    \log
    \frac{\exp(a_{ii})}
    {\sum_{j=1}^{B}\exp(a_{ij})}, \\
    \mathcal{L}_{\mathrm{t2l}}
    &=
    -\frac{1}{B}
    \sum_{i=1}^{B}
    \log
    \frac{\exp(a_{ii})}
    {\sum_{j=1}^{B}\exp(a_{ji})}.
\end{align}
The semantic alignment objective is
\begin{equation}
    \mathcal{L}_{\mathrm{sem}}
    =
    \frac{1}{2}
    \left(
        \mathcal{L}_{\mathrm{l2t}}
        +
        \mathcal{L}_{\mathrm{t2l}}
    \right).
\end{equation}

This objective encourages a coordinate embedding to align with the semantic description of its surrounding urban environment.

\subsection{Distance-Aware Regularisation}
\label{sec:distance_regularisation}

For a mini-batch of $B$ locations, there are $\binom{B}{2}$ possible unordered pairs. We randomly sample $K=\left\lfloor \rho\binom{B}{2} \right\rfloor$ pairs, where $\rho\in(0,1]$ is the pair-sampling ratio. Let $\mathcal{P}_{B}$ denote the sampled pair set.

For each $(i,j)\in\mathcal{P}_{B}$, we calculate the geographic distance $d_{ij} =d_{\mathrm{geo}}(\mathbf{x}_i,\mathbf{x}_j)$. Exact normalisation over all training-location pairs requires $O(N^2)$ pairwise computations and is computationally expensive for datasets containing hundreds of thousands of locations. We therefore apply min--max normalisation within each sampled mini-batch:
\begin{equation}
    \tilde{d}_{ij}
    =
    \frac{
        d_{ij}-d_{\min}^{(B)}
    }{
        d_{\max}^{(B)}-d_{\min}^{(B)}+\epsilon
    },
\end{equation}
where $d_{\min}^{(B)}$ and $d_{\max}^{(B)}$ are the minimum and maximum sampled distances and $\epsilon$ ensures numerical stability.

The cosine similarity between two location embeddings is mapped to $[0,1]$:
\begin{equation}
    \tilde{s}_{ij}
    =
    \frac{
        1+\mathbf{z}_i^{\top}\mathbf{z}_j
    }{2}.
\end{equation}
We define the batch-relative target similarity as
\begin{equation}
    s_{ij}^{*}
    =
    1-\tilde{d}_{ij}
\end{equation}
and minimise
\begin{equation}
    \mathcal{L}_{\mathrm{dist}}
    =
    \frac{1}{K}
    \sum_{(i,j)\in\mathcal{P}_{B}}
    \left(
        \tilde{s}_{ij}-s_{ij}^{*}
    \right)^2.
\end{equation}

The linear target provides a bounded, parameter-free spatial signal. It is used as a soft regulariser rather than an exact geometric constraint: geographically closer locations are encouraged to be more similar, while the repulsive signal increases with geographic separation.

\subsection{Training Objective}
\label{sec:training_objective}

The complete LE4Mob objective is
\begin{equation}
    \mathcal{L}
    =
    \mathcal{L}_{\mathrm{sem}}
    +
    \lambda_{\mathrm{dist}}
    \mathcal{L}_{\mathrm{dist}},
\end{equation}
where $\lambda_{\mathrm{dist}}$ balances semantic alignment and distance-aware regularisation.

During pre-training, the text encoder $g_{\phi}$ is frozen, while the location encoder $f_{\theta}$ and projection layer $p_{\psi}$ are optimised. The resulting location encoder is then frozen and reused across downstream mobility tasks.

\subsection{Integration with Downstream Mobility Models}
\label{sec:downstream_integration}

\subsubsection{Next Location Prediction}

Following previous work \cite{wang2025into,hong2023context}, we use a multi-head self-attention (MHSA) model to process the historical visitation sequence. 
Let $\mathbf{h}^n$ denote its final hidden representation. The next location probabilities are
\begin{equation}
    {\mathbf{P}}(\hat{l}^{n+1})
    =
    \operatorname{Softmax}
    \left(
        \mathrm{MLP}_{\mathrm{o}}(\mathbf{h}^{n})
    \right),
\end{equation}
where $\mathrm{MLP}_{\mathrm{o}}$ denotes the output module consisting of fully-connected layers with residual connections.
The downstream model is trained using standard cross-entropy loss:
\begin{equation}
    \mathcal{L}_\text{next\_loc}
    =
    -
    \sum_{k=1}^{|\mathcal{L}_c|}
    y^{n+1}_{k}
    \log
    {\mathbf{P}}(\hat{l}^{n+1}_k),
\end{equation}
where ${\mathbf{P}}(\hat{l}^{n+1}_k)$ denotes the predicted probability of visiting the $k$th location and $y^{n+1}_{k}$ is the one-hot ground-truth destination. The same MHSA architecture is used for all representation methods, and all pre-trained embeddings remain frozen.

\subsubsection{Commuter Flow Generation}

For each Output Area (OA) $l_i$, we use its population-weighted centroid $\mathbf{x}_i$ as the representative coordinate and obtain its location embedding as $\mathbf{e}_i=f_{\theta}(\mathrm{PE}(\mathbf{x}_i))$.

We evaluate two downstream flow models. The first follows DeepGravity \cite{simini2021deep}, where each OD pair is represented by concatenating the origin embedding, destination embedding, and OD distance: $\mathbf{v}_{ij}
=
[
\mathbf{e}_i;
\mathbf{e}_j;
d_{ij}
]$.

The vector $\mathbf{v}_{ij}$ is passed through a shared feed-forward neural network to produce an OD score $s_{ij}$, which is normalised over candidate destinations:
\begin{equation}
\label{eq:flow_softmax}
\hat{p}_{ij}
=
\frac{
\exp(s_{ij})
}{
\sum_{k \neq i} \exp(s_{ik})
}.
\end{equation}
To assess the dependence on explicit distance, we also evaluate a distance-ablated variant (the results are presented in Section \ref{sec:flow_without_distance}):
$\mathbf{v}_{ij}^{-\mathrm{dist}}
=
[
\mathbf{e}_i;
\mathbf{e}_j
]$. 

The second model is a bilinear decoder, which first computes a bilinear interaction between an OD pair by
\begin{equation}
\mathbf{r}_{ij}
=
\mathbf{e}_i^{\top}
W_{\mathrm{b}}
\mathbf{e}_j,
\end{equation}
where $W_{\mathrm{b}}$ is learnable. The resulting interaction vector is passed through an MLP that outputs one logit for each candidate destination. The destination probabilities are similarly obtained by a Softmax function like Eq. \ref{eq:flow_softmax}.

Both models are trained using the cross entropy between the observed and predicted destination distributions:
\begin{equation}
\mathcal{L}_{\text{flow}}
=
-
\sum_i
\sum_{j \neq i}
\frac{y_{ij}}{O_i}
\log \hat{p}_{ij},
\end{equation}
Estimated flows are obtained from the predicted destination probabilities as defined above.

\section{Experimental Setup}
\label{sec:experiments}

We evaluate LE4Mob on individual-level next location prediction and population-level commuter flow generation. The experiments address three questions:

\begin{itemize}
    \item \textbf{RQ1:} Can our pre-trained location encoder support mobility tasks operating at different scales?
    \item \textbf{RQ2:} How effectively does LE4Mob generalise when previously unseen locations appear after training?
    \item \textbf{RQ3:} Does distance-aware pre-training improve the spatial structure and downstream utility of location embeddings, particularly when distance is not supplied separately to the downstream model?
\end{itemize}

\subsection{Next Location Prediction}
\label{sec:next_location_setup}

\subsubsection{Datasets and Preprocessing}

We use four public mobility datasets: Foursquare New York (FSQ-NYC), Foursquare Tokyo (FSQ-TKY)\cite{yang2014modeling}, Gowalla-London (Gowalla-LD)\cite{cho2011friendship}, and Geolife\cite{zheng2010geolife}. The first three contain location-based social network (LBSN) check-ins, while Geolife contains GNSS trajectories.

FSQ-NYC and FSQ-TKY provide venue information that can be used to construct contextual POI descriptions. Gowalla and Geolife do not provide sufficiently detailed POI context; we therefore obtain POIs for London and Beijing from Foursquare Open Places \cite{fsq2024osp}.

\begin{table*}
  \caption{Basic statistics of the mobility datasets after preprocessing. The mean and standard deviation across users are reported.}
  \label{tab:nlp_data_stats}
  \centering
  \begin{tabular}{lcccc}
    \toprule
     &FSQ-NYC&FSQ-TKY&Gowalla-LD&Geolife\\
    \midrule
    \#Users & 535& 1643& 68& 39\\
    \#Days tracked & 319 & 318& 367& 1703\\
    \#Total unique locations&	4019&	6959&	1065& 1248\\
    \#Stays per user & $178.3\pm206.1$ &	$225.1 \pm 220.9$&	$127.1 \pm 123.3$&	$358.6 \pm 383.6$\\
    \#Stays per user per day& $2.1 \pm 2.1$& $2.7 \pm 2.7$ &	$2.4 \pm 3.0$ &	$2.4 \pm 1.5$\\
    \#Unique locations per user&	$34.7 \pm 21.4$&	$56.6 \pm 33.5$&	$46.3 \pm 44.2$&	$61.5 \pm 52.8$\\
  \bottomrule
\end{tabular}
\end{table*}

For the LBSN datasets, we remove POIs with fewer than ten check-ins and users with fewer than ten records. For Geolife, we remove users with fewer than 50 active days, extract stay points, and cluster spatially proximate stay points into discrete locations. Additional preprocessing details and dataset statistics are reported in Table~\ref{tab:nlp_data_stats}.

\subsubsection{Baselines}

We compare LE4Mob with the following location representation methods:

\begin{itemize}
    \item \textbf{Vanilla-E2E}: a trainable lookup embedding optimised jointly with the downstream predictor;
    \item \textbf{Skip-gram}\cite{mikolov2013efficient}: a Word2Vec-style method that learns location co-occurrence from mobility sequences;
    \item \textbf{POI2Vec}\cite{feng2017poi2vec}: a sequence-based method incorporating spatial proximity through a geographical hierarchy;
    \item \textbf{Geo-Teaser}\cite{zhao2017geo}: a geo-temporal embedding method using spatial and temporal negative sampling;
    \item \textbf{TALE}\cite{wan2021pre}: a time-aware location embedding method based on a temporal tree structure;
    \item \textbf{CaLLiPer} \cite{wang2025into}: the direct predecessor of LE4Mob, trained using semantic alignment without the distance-aware objective.
\end{itemize}


\subsubsection{Conventional Setting}

Each dataset is divided chronologically into non-overlapping training, validation, and test sets using a 60\%--20\%--20\% split. Mobility sequences are constructed using a seven-day sliding observation window, following previous work \cite{hong2023context}.

\subsubsection{Inductive Setting}

The inductive setting simulates deployment in which a small number of previously unseen locations emerge after model training. We first construct the complete location vocabulary and the chronological train--validation--test split. We then randomly select 10\% of locations as a held-out set $\mathcal{L}^{\mathrm{held}}$ and remove every training or validation sequence containing a held-out location. The full test set remains unchanged and therefore contains both seen and unseen locations. To specifically assess inductive generalisation, the inductive results reported in Table \ref{tab:perf_next_loc_pred} are calculated over test samples involving at least one held-out location.

Coordinate-context pairs associated with $\mathcal{L}^{\mathrm{held}}$ are also excluded from LE4Mob and CaLLiPer pre-training. The test set remains unchanged and therefore contains a mixture of seen and unseen locations. 


Each experiment is repeated five times using different random seeds and independently sampled held-out sets. We report the mean and standard deviation.

\subsubsection{Evaluation Metrics}


We evaluate performance from three complementary perspectives: \textbf{exact hit performance}, \textbf{ranking quality}, and \textbf{spatial fidelity}. Exact hit performance is measured using \textbf{Accuracy@$K$}, indicating whether the ground-truth destination appears among the top-$K$ predictions; we report the accuracy metrics at $K \in \{1, 5\}$. Ranking quality is assessed using Normalised Discounted Cumulative Gain at 10 (\textbf{nDCG@10}), which rewards placing the correct destination higher in the ranked list. Spatial fidelity captures how geographically close the predicted locations are to the ground-truth destination, complementing exact-match and ranking-based evaluation. We measure it using Geographic Distance Error at 1 (\textbf{GDE@1}), defined as the geographic distance (in kilometres) between the top-ranked prediction and the ground-truth destination, and Mean Geographic Distance Error at 5 (\textbf{mGDE@5}), defined as the average geographic distance between the top five predictions and the ground-truth destination.

\subsection{Commuter Flow Generation}
\label{sec:flow_setup}

\subsubsection{Study Areas and Data}

We evaluate commuter flow generation in London and Greater Manchester, an international global city and a major metropolitan area in northern England, respectively. Commuting flows and population data at Output Area level are obtained from the UK 2021 Census \cite{ons2023censusod}. POIs are obtained from Foursquare Open Places \cite{fsq2024osp}.

Each OA is represented spatially by its population-weighted centroid. LE4Mob and the other location encoders generate an embedding at this coordinate. 

\subsubsection{Location-Disjoint Split}

For each city, OAs are randomly divided into mutually exclusive training, validation, and test sets using a 65\%--15\%--20\% split.
For each split,
we retain only flows whose origin and destination both belong to that split.
Flows crossing different subsets are excluded. The total outflow is recomputed within each subset.
This node-level inductive split ensures that test OAs and their associated flows are unavailable during downstream training.

\subsubsection{Baselines}

We compare LE4Mob with the following POI-derived urban representations:

\begin{itemize}
    \item \textbf{Hand-Crafted}: Handcrafted feature vectors consisting of POI proportions and population;
    \item \textbf{LDA} \cite{blei2003latent}: a common baseline that learns probabilistic topic representation from regional POI distributions;
    \item \textbf{SPPE}\cite{huang2022sppe}: a POI-based spatial representation method;
    \item \textbf{Urban2Vec}\cite{wang2020urban2vec}: a deep urban representation model learned from geographic context;
    \item \textbf{Space2Vec}\cite{mai2020iclr}: an inductive coordinate-based location encoder; and
    \item \textbf{CaLLiPer}\cite{wang2025multi}: the semantic-focused predecessor of LE4Mob.
\end{itemize}

Hand-Crafted directly contains POI composition and population density, whereas learned baselines use their original pre-trained representations without appending additional socioeconomic variables. 

Some graph learning-based models \cite{liu2020gmel,yin2023convgcnrf} are not included because they generally learn node representations from the full OD graph, violating our location-disjoint evaluation protocol.

\subsubsection{Downstream Flow Models}

Each representation is evaluated with DeepGravity and Bilinear models. 
The standard DeepGravity configuration includes distance and is used in the main performance comparison. To isolate representation-level distance awareness, we additionally remove the distance feature for every representation method. This ablation is analysed separately in Section~\ref{sec:flow_without_distance}. The Bilinear model uses only the origin and destination representations.

\subsubsection{Evaluation Metrics}

The primary metric is the Common Part of Commuters (CPC), also known as the Sørensen--Dice index:
\begin{equation}
    \mathrm{CPC}
    =
    \frac{
        2\sum_{i}\sum_{j\neq i}
        \min(y_{ij},\hat{y}_{ij})
    }{
        \sum_{i}\sum_{j\neq i}y_{ij}
        +
        \sum_{i}\sum_{j\neq i}\hat{y}_{ij}
    }.
\end{equation}
CPC lies in $[0,1]$, with larger values indicating greater overlap between observed and generated flows.

We additionally report Mean Absolute Error (MAE), Root Mean Squared Error (RMSE), and Jensen--Shannon Divergence (JSD) to measure absolute error and distributional dissimilarity.

\subsection{Implementation Details}
\label{sec:implementation}

\subsubsection{Location Representation Pre-training}
For fair comparison, we keep the architectural configuration for the location encoders consistent across Space2Vec, CaLLiPer, and LE4Mob. We employ Grid \cite{mai2020iclr} as the PE and FC-Net as the neural network. The mathematical formulation of Grid, with $\lambda, \phi$ denoting 2-D coordinates, is as follows:

\begin{equation}
    \text{PE}(\lambda,\phi)=\bigcup_{s=0}^{S-1}(\cos \frac{\lambda}{\alpha_s} ,\sin \frac{\lambda}{\alpha_{s}},\cos \frac{\phi}{\alpha_s},\sin \frac{\phi}{\alpha_s})
\end{equation}

\begin{equation}
    \alpha_{s}=r_{\text{min}} \cdot (\frac{r_{\text{max}}}{r_{\text{min}}})^{\frac{s}{S-1}}
\end{equation} where $r_{\text{min}}$ and $r_{\text{max}}$ are the minimum and maximum radii, respectively, and $S$ is the number of scales. These hyperparameters control the resolutions of the multi-scale encoding of the coordinates. We adopted Sentence Transformer \cite{reimers2019sbert} as the text encoder and a linear layer as the projection layer.

The hyperparameters of Grid vary across different cities/datasets: $r_{\text{min}}$ and $r_{\text{max}}$ are set to 0.01 and 10 for New York City (FSQ-NYC) and Tokyo (FSQ-TKY), 1 and 1000 for London (Gowalla-LD and flow generation), 0.01 and 10 for Beijing (Geolife), and 10 and 1000 for Manchester (flow generation). All other hyperparameters remain consistent: we set the number of scales $S=32$ and the hidden dimension of the FC-Net as 256. 

For LE4Mob, the pair-sampling ratio $\rho$ and loss weight $\lambda_{\mathrm{dist}}$ were set as 0.1 and 1, respectively.
For all other baselines, we tuned the hyperparameters via random search and trained them until convergence. 

All location embeddings are set to 128 dimensions except for two baselines in the flow generation task: Hand-Crafted uses 10 dimensions, including 9 POI proportions and 1 population feature, while LDA dimensionality is selected based on perplexity score and set to 14 for London and 8 for Manchester.

All the pre-training has been conducted using a learning rate of 0.001 and the Adam optimiser. Batch sizes vary depending on the volumes of the training data.

\subsubsection{Downstream Models}

The hyperparameters of the MHSA model are kept consistent with the original paper \cite{hong2023context}. For the DeepGravity and Bilinear models, the number of hidden layers is set to 6 and the hidden dimension is set to 128. We tune the dropout rate from {0, 0.2, 0.5}, with the final value varying across models. All downstream models are trained using the Adam optimiser with a learning rate of 0.001. Early stopping is applied. The full settings are provided in the GitHub repository.

To ensure that evaluation results are robust, and that observed improvements are consistent rather than due to random held-out sets or spatial splits, we report the mean and standard deviation over five runs for all tasks. For conventional next location prediction, the five runs use different MHSA initialisation seeds. For inductive next location prediction, they use five differently sampled held-out location sets, with MHSA initialised once. Flow generation task uses five spatially disjoint OA splits generated with different seeds.


\section{Results}

\subsection{Performance on Next Location Prediction}

\setlength{\tabcolsep}{3pt}  
\begin{table*}[t]
\caption{Performance comparison of different embedding methods on the next location prediction task. The best and second-best performance are marked in \textbf{bold} and \underline{underlined}, respectively, based on the unrounded mean values. For better readability, the ranking metric values are scaled by a factor of $10^2$, while GDE and mGDE are reported in kilometres.}
\label{tab:perf_next_loc_pred}
\centering
\begin{tabular}{cl|ccccc|ccccc}
\toprule
\multicolumn{2}{l}{} & \multicolumn{5}{c}{Conventional} & \multicolumn{5}{c}{Inductive} \\
\midrule
 Data & Model & {Acc@1 $\uparrow$} & {Acc@5 $\uparrow$} & {nDCG@10 $\uparrow$} & {GDE@1 $\downarrow$} & {mGDE@5 $\downarrow$} & {Acc@1 $\uparrow$} & {Acc@5 $\uparrow$} & {nDCG@10 $\uparrow$} & {GDE@1 $\downarrow$} & {mGDE@5 $\downarrow$} \\
\midrule
\multirow{7}{*}{\rotatebox{90}{FSQ-NYC}} & Vanilla-E2E & 19.94 {\tiny\ensuremath{\pm}0.19} & 46.84 {\tiny\ensuremath{\pm}0.44} & 37.52 {\tiny\ensuremath{\pm}0.25} & 4.45 {\tiny\ensuremath{\pm}0.09} & 5.36 {\tiny\ensuremath{\pm}0.10} & 9.80 {\tiny\ensuremath{\pm}1.18} & 24.41 {\tiny\ensuremath{\pm}2.10} & 19.23 {\tiny\ensuremath{\pm}1.93} & 7.53 {\tiny\ensuremath{\pm}0.75} & 8.47 {\tiny\ensuremath{\pm}0.69} \\
 & Skip-gram & 19.81 {\tiny\ensuremath{\pm}0.34} & 45.76 {\tiny\ensuremath{\pm}0.19} & 36.85 {\tiny\ensuremath{\pm}0.29} & 4.43 {\tiny\ensuremath{\pm}0.06} & 5.20 {\tiny\ensuremath{\pm}0.03} & 8.68 {\tiny\ensuremath{\pm}1.19} & 21.71 {\tiny\ensuremath{\pm}2.92} & 17.35 {\tiny\ensuremath{\pm}2.40} & 8.07 {\tiny\ensuremath{\pm}0.82} & 8.96 {\tiny\ensuremath{\pm}1.15} \\
 & POI2Vec & 19.62 {\tiny\ensuremath{\pm}0.58} & 44.94 {\tiny\ensuremath{\pm}0.71} & 36.32 {\tiny\ensuremath{\pm}0.66} & 4.53 {\tiny\ensuremath{\pm}0.16} & 5.38 {\tiny\ensuremath{\pm}0.12} & 8.84 {\tiny\ensuremath{\pm}1.31} & 21.80 {\tiny\ensuremath{\pm}3.40} & 17.37 {\tiny\ensuremath{\pm}2.70} & 8.26 {\tiny\ensuremath{\pm}0.73} & 8.74 {\tiny\ensuremath{\pm}0.72} \\
 & Geo-Teaser & 19.43 {\tiny\ensuremath{\pm}0.73} & 45.46 {\tiny\ensuremath{\pm}0.84} & 36.48 {\tiny\ensuremath{\pm}0.59} & 4.52 {\tiny\ensuremath{\pm}0.08} & 5.30 {\tiny\ensuremath{\pm}0.08} & 8.08 {\tiny\ensuremath{\pm}1.47} & 20.33 {\tiny\ensuremath{\pm}3.13} & 16.16 {\tiny\ensuremath{\pm}2.74} & 8.77 {\tiny\ensuremath{\pm}0.48} & 9.26 {\tiny\ensuremath{\pm}0.61} \\
 & TALE & 20.20 {\tiny\ensuremath{\pm}0.40} & 45.80 {\tiny\ensuremath{\pm}0.12} & 36.98 {\tiny\ensuremath{\pm}0.12} & 4.46 {\tiny\ensuremath{\pm}0.06} & 5.35 {\tiny\ensuremath{\pm}0.05} & 8.52 {\tiny\ensuremath{\pm}1.47} & 21.04 {\tiny\ensuremath{\pm}3.41} & 16.83 {\tiny\ensuremath{\pm}2.76} & 8.23 {\tiny\ensuremath{\pm}0.99} & 9.07 {\tiny\ensuremath{\pm}0.99} \\
 & CaLLiPer & \underline{20.33 {\tiny\ensuremath{\pm}0.36}} & \underline{48.38 {\tiny\ensuremath{\pm}0.15}} & \underline{38.67 {\tiny\ensuremath{\pm}0.12}} & \underline{4.10 {\tiny\ensuremath{\pm}0.03}} & \underline{4.85 {\tiny\ensuremath{\pm}0.02}} & \underline{10.63 {\tiny\ensuremath{\pm}0.97}} & \underline{26.59 {\tiny\ensuremath{\pm}1.69}} & \underline{21.13 {\tiny\ensuremath{\pm}1.56}} & \underline{5.44 {\tiny\ensuremath{\pm}0.35}} & \underline{5.97 {\tiny\ensuremath{\pm}0.37}} \\
 & LE4Mob & \textbf{20.45 {\tiny\ensuremath{\pm}0.32}} & \textbf{48.43 {\tiny\ensuremath{\pm}0.36}} & \textbf{38.73 {\tiny\ensuremath{\pm}0.18}} & \textbf{4.09 {\tiny\ensuremath{\pm}0.05}} & \textbf{4.82 {\tiny\ensuremath{\pm}0.02}} & \textbf{10.74 {\tiny\ensuremath{\pm}1.33}} & \textbf{27.04 {\tiny\ensuremath{\pm}2.40}} & \textbf{21.37 {\tiny\ensuremath{\pm}2.09}} & \textbf{5.20 {\tiny\ensuremath{\pm}0.39}} & \textbf{5.65 {\tiny\ensuremath{\pm}0.27}} \\

\midrule
\multirow{7}{*}{\rotatebox{90}{FSQ-TKY}} & Vanilla-E2E & 21.54 {\tiny\ensuremath{\pm}0.06} & 45.54 {\tiny\ensuremath{\pm}0.16} & 37.32 {\tiny\ensuremath{\pm}0.03} & 4.58 {\tiny\ensuremath{\pm}0.02} & 5.41 {\tiny\ensuremath{\pm}0.01} & 11.72 {\tiny\ensuremath{\pm}0.48} & 28.01 {\tiny\ensuremath{\pm}1.04} & 22.45 {\tiny\ensuremath{\pm}0.93} & 5.86 {\tiny\ensuremath{\pm}0.25} & 6.56 {\tiny\ensuremath{\pm}0.22} \\
 & Skip-gram & 21.85 {\tiny\ensuremath{\pm}0.23} & \underline{46.38 {\tiny\ensuremath{\pm}0.10}} & 37.93 {\tiny\ensuremath{\pm}0.13} & 4.41 {\tiny\ensuremath{\pm}0.03} & 5.22 {\tiny\ensuremath{\pm}0.02} & 12.02 {\tiny\ensuremath{\pm}0.59} & 28.74 {\tiny\ensuremath{\pm}1.12} & 23.07 {\tiny\ensuremath{\pm}1.01} & 5.49 {\tiny\ensuremath{\pm}0.19} & 6.09 {\tiny\ensuremath{\pm}0.16} \\
 & POI2Vec & \textbf{22.03 {\tiny\ensuremath{\pm}0.13}} & 46.35 {\tiny\ensuremath{\pm}0.07} & \underline{37.98 {\tiny\ensuremath{\pm}0.06}} & \textbf{4.40 {\tiny\ensuremath{\pm}0.02}} & \underline{5.21 {\tiny\ensuremath{\pm}0.00}} & \textbf{12.18 {\tiny\ensuremath{\pm}0.74}} & 28.71 {\tiny\ensuremath{\pm}1.50} & 23.08 {\tiny\ensuremath{\pm}1.20} & 5.63 {\tiny\ensuremath{\pm}0.31} & 6.26 {\tiny\ensuremath{\pm}0.29} \\
 & Geo-Teaser & \underline{21.86 {\tiny\ensuremath{\pm}0.19}} & \textbf{46.70 {\tiny\ensuremath{\pm}0.14}} & \textbf{38.13 {\tiny\ensuremath{\pm}0.14}} & \underline{4.40 {\tiny\ensuremath{\pm}0.02}} & \textbf{5.21 {\tiny\ensuremath{\pm}0.02}} & 11.93 {\tiny\ensuremath{\pm}0.70} & 28.56 {\tiny\ensuremath{\pm}1.29} & 22.93 {\tiny\ensuremath{\pm}1.12} & 5.55 {\tiny\ensuremath{\pm}0.27} & 6.15 {\tiny\ensuremath{\pm}0.20} \\
 & TALE & 21.44 {\tiny\ensuremath{\pm}0.09} & 45.83 {\tiny\ensuremath{\pm}0.03} & 37.46 {\tiny\ensuremath{\pm}0.06} & 4.60 {\tiny\ensuremath{\pm}0.01} & 5.42 {\tiny\ensuremath{\pm}0.02} & 11.92 {\tiny\ensuremath{\pm}0.51} & 28.78 {\tiny\ensuremath{\pm}1.19} & 22.96 {\tiny\ensuremath{\pm}0.95} & 5.81 {\tiny\ensuremath{\pm}0.28} & 6.47 {\tiny\ensuremath{\pm}0.24} \\
 & CaLLiPer & 20.20 {\tiny\ensuremath{\pm}0.14} & 46.19 {\tiny\ensuremath{\pm}0.09} & 37.19 {\tiny\ensuremath{\pm}0.07} & 4.51 {\tiny\ensuremath{\pm}0.02} & 5.25 {\tiny\ensuremath{\pm}0.01} & \underline{12.07 {\tiny\ensuremath{\pm}0.45}} & \underline{29.72 {\tiny\ensuremath{\pm}1.28}} & \underline{23.69 {\tiny\ensuremath{\pm}0.97}} & \underline{5.37 {\tiny\ensuremath{\pm}0.24}} & \underline{5.94 {\tiny\ensuremath{\pm}0.23}} \\
 & LE4Mob & 20.26 {\tiny\ensuremath{\pm}0.11} & 46.33 {\tiny\ensuremath{\pm}0.09} & 37.29 {\tiny\ensuremath{\pm}0.06} & 4.47 {\tiny\ensuremath{\pm}0.02} & 5.22 {\tiny\ensuremath{\pm}0.01} & 12.04 {\tiny\ensuremath{\pm}0.48} & \textbf{29.88 {\tiny\ensuremath{\pm}1.33}} & \textbf{23.77 {\tiny\ensuremath{\pm}0.96}} & \textbf{5.35 {\tiny\ensuremath{\pm}0.26}} & \textbf{5.84 {\tiny\ensuremath{\pm}0.21}} \\

\midrule
\multirow{7}{*}{\rotatebox{90}{Gowalla-LD}} & Vanilla-E2E & 13.19 {\tiny\ensuremath{\pm}0.98} & 28.05 {\tiny\ensuremath{\pm}2.14} & 22.54 {\tiny\ensuremath{\pm}1.43} & 6.04 {\tiny\ensuremath{\pm}0.29} & 7.13 {\tiny\ensuremath{\pm}0.43} & 5.18 {\tiny\ensuremath{\pm}2.13} & 10.79 {\tiny\ensuremath{\pm}4.94} & 9.10 {\tiny\ensuremath{\pm}3.97} & 8.62 {\tiny\ensuremath{\pm}0.72} & 9.03 {\tiny\ensuremath{\pm}0.81} \\
 & Skip-gram & \underline{16.67 {\tiny\ensuremath{\pm}0.77}} & 35.67 {\tiny\ensuremath{\pm}1.36} & 28.85 {\tiny\ensuremath{\pm}0.70} & 4.79 {\tiny\ensuremath{\pm}0.36} & 6.03 {\tiny\ensuremath{\pm}0.22} & \underline{8.65 {\tiny\ensuremath{\pm}3.53}} & \underline{18.18 {\tiny\ensuremath{\pm}6.23}} & \underline{14.55 {\tiny\ensuremath{\pm}4.72}} & \textbf{6.73 {\tiny\ensuremath{\pm}1.02}} & \underline{7.59 {\tiny\ensuremath{\pm}0.61}} \\
 & POI2Vec & 15.67 {\tiny\ensuremath{\pm}0.89} & 33.06 {\tiny\ensuremath{\pm}1.49} & 27.09 {\tiny\ensuremath{\pm}1.26} & 5.18 {\tiny\ensuremath{\pm}0.15} & 6.15 {\tiny\ensuremath{\pm}0.16} & 7.36 {\tiny\ensuremath{\pm}2.19} & 16.29 {\tiny\ensuremath{\pm}5.00} & 13.36 {\tiny\ensuremath{\pm}3.93} & 8.20 {\tiny\ensuremath{\pm}1.03} & 8.80 {\tiny\ensuremath{\pm}0.95} \\
 & Geo-Teaser & 16.48 {\tiny\ensuremath{\pm}0.61} & 35.20 {\tiny\ensuremath{\pm}0.44} & 28.38 {\tiny\ensuremath{\pm}0.35} & 4.64 {\tiny\ensuremath{\pm}0.29} & 5.82 {\tiny\ensuremath{\pm}0.27} & 8.27 {\tiny\ensuremath{\pm}3.09} & 17.41 {\tiny\ensuremath{\pm}6.47} & 14.05 {\tiny\ensuremath{\pm}4.71} & 7.67 {\tiny\ensuremath{\pm}2.30} & 7.80 {\tiny\ensuremath{\pm}1.19} \\
 & TALE & 14.73 {\tiny\ensuremath{\pm}0.97} & 30.33 {\tiny\ensuremath{\pm}1.65} & 24.67 {\tiny\ensuremath{\pm}0.95} & 5.56 {\tiny\ensuremath{\pm}0.39} & 6.76 {\tiny\ensuremath{\pm}0.24} & 6.95 {\tiny\ensuremath{\pm}3.27} & 15.44 {\tiny\ensuremath{\pm}4.97} & 12.40 {\tiny\ensuremath{\pm}4.22} & 7.08 {\tiny\ensuremath{\pm}0.91} & 7.97 {\tiny\ensuremath{\pm}0.87} \\
 & CaLLiPer & 16.58 {\tiny\ensuremath{\pm}1.09} & \underline{36.07 {\tiny\ensuremath{\pm}2.18}} & \underline{29.35 {\tiny\ensuremath{\pm}1.35}} & \underline{4.45 {\tiny\ensuremath{\pm}0.25}} & \underline{5.62 {\tiny\ensuremath{\pm}0.21}} & 5.77 {\tiny\ensuremath{\pm}2.77} & 14.33 {\tiny\ensuremath{\pm}3.56} & 11.47 {\tiny\ensuremath{\pm}3.54} & 7.83 {\tiny\ensuremath{\pm}1.35} & 8.15 {\tiny\ensuremath{\pm}0.80} \\
 & LE4Mob & \textbf{19.02 {\tiny\ensuremath{\pm}0.55}} & \textbf{39.34 {\tiny\ensuremath{\pm}1.14}} & \textbf{31.96 {\tiny\ensuremath{\pm}0.37}} & \textbf{4.39 {\tiny\ensuremath{\pm}0.12}} & \textbf{5.55 {\tiny\ensuremath{\pm}0.32}} & \textbf{9.80 {\tiny\ensuremath{\pm}3.85}} & \textbf{22.79 {\tiny\ensuremath{\pm}8.25}} & \textbf{17.90 {\tiny\ensuremath{\pm}6.27}} & \underline{6.90 {\tiny\ensuremath{\pm}2.17}} & \textbf{7.54 {\tiny\ensuremath{\pm}1.34}} \\

\midrule
\multirow{7}{*}{\rotatebox{90}{Geolife}} & Vanilla-E2E & 44.15 {\tiny\ensuremath{\pm}0.54} & 74.72 {\tiny\ensuremath{\pm}2.32} & 62.79 {\tiny\ensuremath{\pm}0.86} & 2.71 {\tiny\ensuremath{\pm}0.05} & 4.19 {\tiny\ensuremath{\pm}0.14} & 32.16 {\tiny\ensuremath{\pm}3.34} & 53.27 {\tiny\ensuremath{\pm}6.01} & 44.66 {\tiny\ensuremath{\pm}4.73} & 3.86 {\tiny\ensuremath{\pm}0.84} & 4.93 {\tiny\ensuremath{\pm}1.00} \\
 & Skip-gram & 45.32 {\tiny\ensuremath{\pm}0.50} & \textbf{77.75 {\tiny\ensuremath{\pm}1.80}} & \underline{64.36 {\tiny\ensuremath{\pm}0.94}} & 2.84 {\tiny\ensuremath{\pm}0.07} & 3.92 {\tiny\ensuremath{\pm}0.08} & 33.10 {\tiny\ensuremath{\pm}4.94} & \underline{56.19 {\tiny\ensuremath{\pm}6.08}} & 46.35 {\tiny\ensuremath{\pm}5.92} & 3.67 {\tiny\ensuremath{\pm}0.81} & 4.67 {\tiny\ensuremath{\pm}1.08} \\
 & POI2Vec & 43.08 {\tiny\ensuremath{\pm}1.51} & 74.53 {\tiny\ensuremath{\pm}1.88} & 62.17 {\tiny\ensuremath{\pm}1.18} & 2.82 {\tiny\ensuremath{\pm}0.07} & 3.93 {\tiny\ensuremath{\pm}0.03} & 29.71 {\tiny\ensuremath{\pm}5.10} & 51.11 {\tiny\ensuremath{\pm}7.02} & 42.34 {\tiny\ensuremath{\pm}6.27} & 3.84 {\tiny\ensuremath{\pm}0.93} & 5.17 {\tiny\ensuremath{\pm}0.96} \\
 & Geo-Teaser & \underline{46.30 {\tiny\ensuremath{\pm}1.12}} & 75.01 {\tiny\ensuremath{\pm}0.95} & 63.50 {\tiny\ensuremath{\pm}0.77} & 2.74 {\tiny\ensuremath{\pm}0.10} & \underline{3.88 {\tiny\ensuremath{\pm}0.05}} & {34.91 {\tiny\ensuremath{\pm}4.43}} & 55.53 {\tiny\ensuremath{\pm}7.71} & \underline{47.51 {\tiny\ensuremath{\pm}5.98}} & 3.54 {\tiny\ensuremath{\pm}0.96} & 4.64 {\tiny\ensuremath{\pm}1.13} \\
 & TALE & 45.09 {\tiny\ensuremath{\pm}1.87} & 74.75 {\tiny\ensuremath{\pm}2.62} & 62.92 {\tiny\ensuremath{\pm}1.89} & 2.82 {\tiny\ensuremath{\pm}0.11} & 3.99 {\tiny\ensuremath{\pm}0.18} & 34.92 {\tiny\ensuremath{\pm}6.24} & 55.81 {\tiny\ensuremath{\pm}7.24} & 47.08 {\tiny\ensuremath{\pm}7.04} & 3.61 {\tiny\ensuremath{\pm}0.87} & 4.69 {\tiny\ensuremath{\pm}1.01} \\
 & CaLLiPer & 46.28 {\tiny\ensuremath{\pm}2.62} & 74.36 {\tiny\ensuremath{\pm}0.45} & 63.22 {\tiny\ensuremath{\pm}1.69} & \underline{2.64 {\tiny\ensuremath{\pm}0.15}} & 3.90 {\tiny\ensuremath{\pm}0.07} & \underline{35.18 {\tiny\ensuremath{\pm}4.05}} & \textbf{56.34 {\tiny\ensuremath{\pm}7.32}} & \textbf{47.63 {\tiny\ensuremath{\pm}5.71}} & \underline{3.51 {\tiny\ensuremath{\pm}1.09}} & \underline{4.60 {\tiny\ensuremath{\pm}1.05}} \\
 & LE4Mob & \textbf{46.55 {\tiny\ensuremath{\pm}1.30}} & \underline{76.43 {\tiny\ensuremath{\pm}0.77}} & \textbf{64.54 {\tiny\ensuremath{\pm}0.32}} & \textbf{2.57 {\tiny\ensuremath{\pm}0.12}} & \textbf{3.83 {\tiny\ensuremath{\pm}0.15}} & \textbf{35.35 {\tiny\ensuremath{\pm}4.30}} & 55.34 {\tiny\ensuremath{\pm}8.46} & 47.07 {\tiny\ensuremath{\pm}6.41} & \textbf{3.39 {\tiny\ensuremath{\pm}0.87}} & \textbf{4.50 {\tiny\ensuremath{\pm}0.92}} \\

\bottomrule
\end{tabular}
\end{table*}


Table \ref{tab:perf_next_loc_pred} compares the location representations under conventional and inductive next location prediction. \textbf{LE4Mob achieves the highest mean performance in 30 of the 40 dataset–setting–metric combinations, including 13 of the 16 spatial error comparisons}. On FSQ-NYC, it consistently outperforms all baselines across both ranking and spatial metrics under conventional and inductive evaluation. Strong improvements are also observed on Gowalla-LD: across the three ranking metrics, LE4Mob improves over the strongest competing method by approximately 8.9–14.1\% under conventional evaluation and 13.3–25.4\% under inductive evaluation. Its spatial errors are also generally favourable. On FSQ-TKY, trajectory-derived representations remain stronger under conventional evaluation, whereas LE4Mob achieves the best performance on four of the five inductive metrics, with POI2Vec leading Acc@1. On Geolife, LE4Mob achieves the best results in seven of the ten reported comparisons, including all four spatial error metrics.
Overall, LE4Mob shows its most consistent advantages under inductive evaluation and on metrics that directly measure the spatial proximity of predictions.

Baseline performance varies across datasets and evaluation settings. Trajectory-derived embeddings perform well when their learned mobility patterns match a dataset, such as POI2Vec and Geo-Teaser on FSQ-TKY, and Skip-gram on Gowalla-LD and Geolife. 
CaLLiPer is the most consistently competitive baseline, especially on FSQ-NYC and Geolife, highlighting the value of inductive spatial-semantic representations. In comparison, LE4Mob achieves the best overall results across substantially more dataset–metric combinations, demonstrating greater consistency across heterogeneous mobility environments. It also records the lowest error in 13 of the 16 spatial error comparisons, indicating that distance-aware pre-training improves both destination ranking and the spatial proximity of predictions.

\subsection{Performance on Flow Generation}

\setlength{\tabcolsep}{6pt}  
\begin{table*}[t]
\caption{Performance comparison of different embedding methods on the flow generation task. The best and second-best performances are determined using unrounded mean values and marked in \textbf{bold} and \underline{underlined}, respectively.}
\label{tab:perf_flow_generation}
\centering
\begin{tabular}{cl|cccc|cccc}
\toprule
\multicolumn{2}{l}{} &
\multicolumn{4}{c}{DeepGravity} &
\multicolumn{4}{c}{Bilinear} \\
\midrule
Data & Model &
{CPC $\uparrow$} &
{MAE $\downarrow$} &
{RMSE $\downarrow$} &
{JSD $\downarrow$} &
{CPC $\uparrow$} &
{MAE $\downarrow$} &
{RMSE $\downarrow$} &
{JSD $\downarrow$} \\
\midrule

\multirow{8}{*}{\rotatebox{90}{Manchester}}
& Handcrafted
& 0.765 {\tiny\ensuremath{\pm}0.008}
& 0.693 {\tiny\ensuremath{\pm}0.055}
& 0.871 {\tiny\ensuremath{\pm}0.083}
& 0.202 {\tiny\ensuremath{\pm}0.005}
& 0.757 {\tiny\ensuremath{\pm}0.011}
& 0.717 {\tiny\ensuremath{\pm}0.063}
& 0.915 {\tiny\ensuremath{\pm}0.091}
& 0.207 {\tiny\ensuremath{\pm}0.008} \\


& LDA
& \textbf{0.784 {\tiny\ensuremath{\pm}0.017}}
& \textbf{0.649 {\tiny\ensuremath{\pm}0.080}}
& \textbf{0.835 {\tiny\ensuremath{\pm}0.116}}
& \textbf{0.183 {\tiny\ensuremath{\pm}0.013}}
& 0.769 {\tiny\ensuremath{\pm}0.021}
& 0.694 {\tiny\ensuremath{\pm}0.095}
& 0.904 {\tiny\ensuremath{\pm}0.142}
& 0.192 {\tiny\ensuremath{\pm}0.019} \\

& SPPE
& 0.778 {\tiny\ensuremath{\pm}0.012}
& 0.669 {\tiny\ensuremath{\pm}0.066}
& 0.866 {\tiny\ensuremath{\pm}0.097}
& 0.189 {\tiny\ensuremath{\pm}0.009}
& 0.796 {\tiny\ensuremath{\pm}0.009}
& 0.610 {\tiny\ensuremath{\pm}0.058}
& 0.795 {\tiny\ensuremath{\pm}0.089}
& 0.173 {\tiny\ensuremath{\pm}0.006} \\

& Urban2Vec
& 0.769 {\tiny\ensuremath{\pm}0.017}
& 0.693 {\tiny\ensuremath{\pm}0.083}
& 0.895 {\tiny\ensuremath{\pm}0.119}
& 0.196 {\tiny\ensuremath{\pm}0.014}
& \underline{0.809 {\tiny\ensuremath{\pm}0.017}}
& \underline{0.584 {\tiny\ensuremath{\pm}0.078}}
& \underline{0.793 {\tiny\ensuremath{\pm}0.109}}
& \underline{0.157 {\tiny\ensuremath{\pm}0.012}} \\

& Space2Vec
& 0.764 {\tiny\ensuremath{\pm}0.016}
& 0.705 {\tiny\ensuremath{\pm}0.080}
& 0.903 {\tiny\ensuremath{\pm}0.116}
& 0.199 {\tiny\ensuremath{\pm}0.013}
& 0.783 {\tiny\ensuremath{\pm}0.022}
& 0.660 {\tiny\ensuremath{\pm}0.095}
& 0.892 {\tiny\ensuremath{\pm}0.140}
& 0.183 {\tiny\ensuremath{\pm}0.019} \\

& CaLLiPer
& 0.775 {\tiny\ensuremath{\pm}0.015}
& 0.675 {\tiny\ensuremath{\pm}0.074}
& 0.869 {\tiny\ensuremath{\pm}0.108}
& 0.188 {\tiny\ensuremath{\pm}0.012}
& 0.806 {\tiny\ensuremath{\pm}0.017}
& 0.603 {\tiny\ensuremath{\pm}0.080}
& 0.821 {\tiny\ensuremath{\pm}0.121}
& 0.163 {\tiny\ensuremath{\pm}0.015} \\

& LE4Mob
& \underline{0.779 {\tiny\ensuremath{\pm}0.014}}
& \underline{0.664 {\tiny\ensuremath{\pm}0.073}}
& \underline{0.858 {\tiny\ensuremath{\pm}0.107}}
& \underline{0.184 {\tiny\ensuremath{\pm}0.012}}
& \textbf{0.814 {\tiny\ensuremath{\pm}0.015}}
& \textbf{0.582 {\tiny\ensuremath{\pm}0.072}}
& \textbf{0.788 {\tiny\ensuremath{\pm}0.107}}
& \textbf{0.154 {\tiny\ensuremath{\pm}0.012}} \\

\midrule

\multirow{8}{*}{\rotatebox{90}{London}}
& Handcrafted
& 0.786 {\tiny\ensuremath{\pm}0.002}
& 0.579 {\tiny\ensuremath{\pm}0.022}
& 0.715 {\tiny\ensuremath{\pm}0.038}
& 0.186 {\tiny\ensuremath{\pm}0.002}
& 0.800 {\tiny\ensuremath{\pm}0.005}
& 0.542 {\tiny\ensuremath{\pm}0.025}
& 0.686 {\tiny\ensuremath{\pm}0.044}
& 0.175 {\tiny\ensuremath{\pm}0.005} \\


& LDA
& 0.809 {\tiny\ensuremath{\pm}0.003}
& 0.524 {\tiny\ensuremath{\pm}0.021}
& 0.662 {\tiny\ensuremath{\pm}0.035}
& 0.165 {\tiny\ensuremath{\pm}0.002}
& 0.797 {\tiny\ensuremath{\pm}0.007}
& 0.555 {\tiny\ensuremath{\pm}0.032}
& 0.710 {\tiny\ensuremath{\pm}0.052}
& 0.171 {\tiny\ensuremath{\pm}0.006} \\

& SPPE
& \underline{0.812 {\tiny\ensuremath{\pm}0.004}}
& \underline{0.516 {\tiny\ensuremath{\pm}0.025}}
& \underline{0.651 {\tiny\ensuremath{\pm}0.041}}
& \underline{0.160 {\tiny\ensuremath{\pm}0.003}}
& 0.810 {\tiny\ensuremath{\pm}0.003}
& 0.523 {\tiny\ensuremath{\pm}0.022}
& 0.666 {\tiny\ensuremath{\pm}0.040}
& 0.159 {\tiny\ensuremath{\pm}0.003} \\

& Urban2Vec
& 0.793 {\tiny\ensuremath{\pm}0.008}
& 0.564 {\tiny\ensuremath{\pm}0.035}
& 0.707 {\tiny\ensuremath{\pm}0.053}
& 0.175 {\tiny\ensuremath{\pm}0.007}
& \underline{0.826 {\tiny\ensuremath{\pm}0.007}}
& \underline{0.487 {\tiny\ensuremath{\pm}0.027}}
& 0.676 {\tiny\ensuremath{\pm}0.049}
& \underline{0.150 {\tiny\ensuremath{\pm}0.007}} \\

& Space2Vec
& 0.798 {\tiny\ensuremath{\pm}0.008}
& 0.554 {\tiny\ensuremath{\pm}0.036}
& 0.689 {\tiny\ensuremath{\pm}0.054}
& 0.173 {\tiny\ensuremath{\pm}0.007}
& 0.813 {\tiny\ensuremath{\pm}0.008}
& 0.516 {\tiny\ensuremath{\pm}0.035}
& \underline{0.651 {\tiny\ensuremath{\pm}0.053}}
& 0.157 {\tiny\ensuremath{\pm}0.007} \\

& CaLLiPer
& 0.807 {\tiny\ensuremath{\pm}0.008}
& 0.532 {\tiny\ensuremath{\pm}0.035}
& 0.671 {\tiny\ensuremath{\pm}0.053}
& 0.162 {\tiny\ensuremath{\pm}0.007}
& 0.817 {\tiny\ensuremath{\pm}0.008}
& 0.508 {\tiny\ensuremath{\pm}0.032}
& 0.671 {\tiny\ensuremath{\pm}0.054}
& 0.153 {\tiny\ensuremath{\pm}0.007} \\

& LE4Mob
& \textbf{0.830 {\tiny\ensuremath{\pm}0.010}}
& \textbf{0.477 {\tiny\ensuremath{\pm}0.039}}
& \textbf{0.630 {\tiny\ensuremath{\pm}0.060}}
& \textbf{0.142 {\tiny\ensuremath{\pm}0.009}}
& \textbf{0.838 {\tiny\ensuremath{\pm}0.010}}
& \textbf{0.459 {\tiny\ensuremath{\pm}0.037}}
& \textbf{0.613 {\tiny\ensuremath{\pm}0.058}}
& \textbf{0.135 {\tiny\ensuremath{\pm}0.008}} \\

\bottomrule
\end{tabular}
\end{table*}


Table~\ref{tab:perf_flow_generation} compares the embedding methods using DeepGravity and Bilinear downstream models. Among all embedding--model combinations, \textbf{LE4Mob with Bilinear achieves the strongest overall results}, while LE4Mob records the highest individual mean performance in \textbf{12 of the 16 city--model--metric combinations}. It also consistently outperforms CaLLiPer across all evaluated configurations, indicating that explicitly preserving spatial proximity improves the utility of spatial-semantic representations for flow generation.


In Greater Manchester, LE4Mob obtains the strongest Bilinear results across all four metrics. The DeepGravity results present the main exception: LDA performs best, while LE4Mob remains competitive, and continues to outperform CaLLiPer. This suggests that the effectiveness of a location representation depends partly on how the downstream architecture incorporates spatial information. As DeepGravity already receives explicit OD distance, the spatial signal encoded by LE4Mob may become partly redundant in this configuration.

In London, LE4Mob consistently outperforms all baseline methods under both downstream architectures. With DeepGravity, it improves CPC by 2.2\% and reduces MAE, RMSE, and JSD by 7.6\%, 3.2\%, and 11.3\%, respectively, relative to the strongest competing method. With Bilinear, it improves CPC by 1.5\% and reduces MAE, RMSE, and JSD by 5.7\%, 5.8\%, and 10.0\%, respectively, compared to the second-best results. These results demonstrate that LE4Mob can support both nonlinear feature-based and direct origin--destination interaction models

The baseline methods exhibit different strengths across configurations: LDA perform strongly with DeepGravity in Greater Manchester, SPPE is competitive with DeepGravity in London, and Urban2Vec is the strongest Bilinear baseline. However, none of these methods performs better consistently across all configurations. In comparison, LE4Mob provides the strongest overall performance and the most consistent improvements across cities, downstream models, and evaluation metrics.

\subsection{Analysis of Distance Awareness}

\subsubsection{Correspondence between Geographic and Embedding Similarities}

\begin{table}[t]
\caption{Correspondence between geographic similarity and embedding cosine
similarity for OA pairs in the Greater London test set. All correlations are
statistically significant at $p<0.0001$.}
\label{tab:distance_correlation}
\centering
\begin{tabular}{lcc}
\toprule
Model & Pearson's $r$ & Spearman's $\rho$ \\
\midrule
CaLLiPer & 0.268 & 0.315 \\
LE4Mob   & \textbf{0.369} & \textbf{0.438} \\
\bottomrule
\end{tabular}
\end{table}

We next examine whether the distance-preserving objective successfully incorporates geographic proximity into the learned location embeddings. We compare LE4Mob with CaLLiPer, as LE4Mob extends the same spatial-semantic representation framework by introducing distance-aware regularisation. London is selected as the case-study area.

For each pair of OAs in the test set, we define geographic similarity as
$s^{\mathrm{geo}}_{ij}=1-\tilde{d}_{ij}$, where $\tilde{d}_{ij}$ is the
min-max normalised Euclidean distance between their centroids. Embedding
similarity is measured using the cosine similarity between the corresponding
embedding vectors. We then calculate Pearson's correlation coefficient to
assess the linear correspondence between geographic and embedding similarities,
and Spearman's rank correlation coefficient to assess the preservation of their
relative ordering. Higher coefficient values indicate
stronger preservation of geographic proximity.

As shown in Table~\ref{tab:distance_correlation}, both methods exhibit statistically significant positive correlations. However, LE4Mob increases the Pearson correlation from 0.268 to 0.369 and the Spearman correlation from 0.315 to 0.438, representing relative improvements of approximately 38\% and 39\%, respectively. The higher Pearson correlation indicates that embedding similarity varies more consistently with the magnitude of geographic proximity, while the higher Spearman correlation shows that LE4Mob better preserves the relative ordering of nearby and distant location pairs. These results provide direct evidence that the distance-preserving objective more effectively incorporates physical spatial relationships into the learned embeddings.

\subsubsection{Flow Generation without Explicit Distance Inputs}
\label{sec:flow_without_distance}
\begin{figure*}[t]
    \centering
    \includegraphics[width=0.9\textwidth]{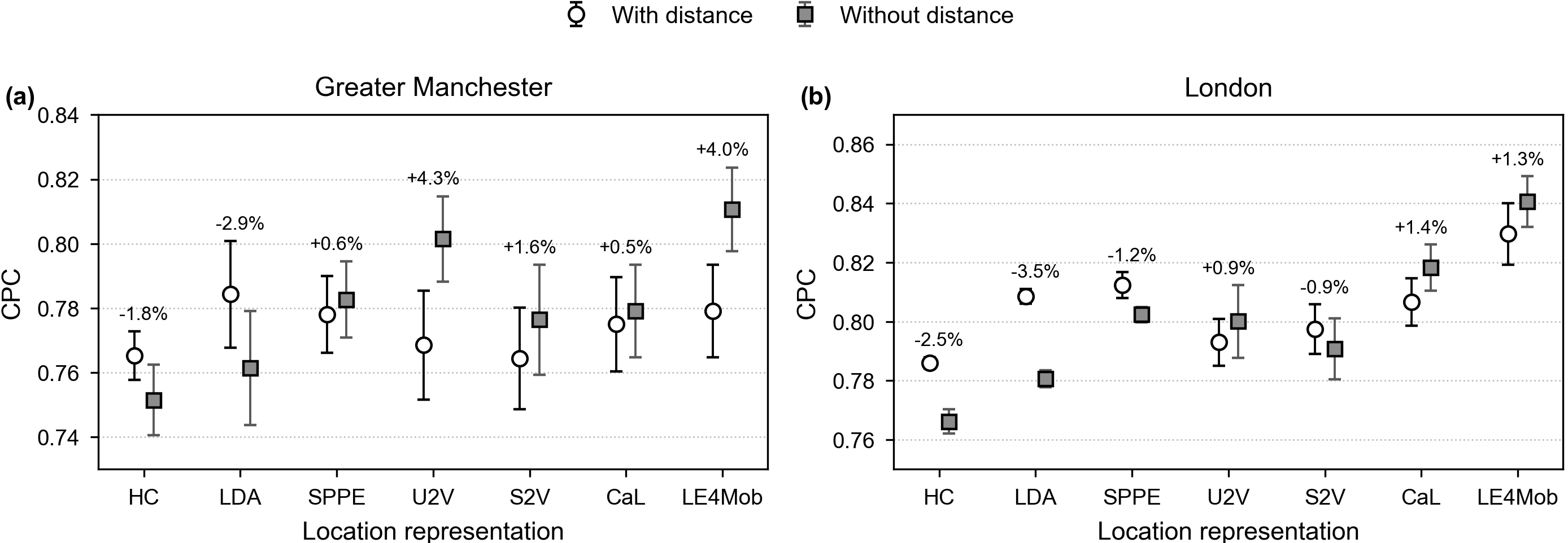}
    \caption{Flow generation performance with and without explicit distance inputs in the downstream DeepGravity model. Error bars indicate standard deviations across repeated runs. Percentages denote the relative CPC change after removing the explicit distance feature. HC, U2V, S2V, and CaL denote Hand-Crafted, Urban2Vec, Space2Vec, and CaLLiPer, respectively.}
    \label{fig:flow_without_distance}
\end{figure*}

To examine whether distance information is encoded directly in the location representations, we remove the explicit origin--destination distance feature from DeepGravity and compare the resulting CPC with the standard configuration (the left part of Table \ref{tab:perf_flow_generation}). As shown in Figure~\ref{fig:flow_without_distance}, removing distance consistently reduces the performance of Hand-Crafted and LDA representations. Their CPC decreases by \(1.80\%\)--\(2.91\%\) in Greater Manchester and \(2.52\%\)--\(3.45\%\) in London. This is because that these representations do not explicitly preserve the spatial relationships between regions and the downstream DeepGravity model rely substantially on the separately provided distance feature.

The downstream DeepGravity model generally produces more robust results after the removal of distance when paired with deep location embeddings, although the responses vary across methods and cities. Most notably, LE4Mob improves from 0.7791 to 0.8106 in Greater Manchester and from 0.8297 to 0.8406 in London, achieving the highest CPC among all representations in both cities when distance is excluded. CaLLiPer also remains robust, with improvements of \(0.53\%\) and \(1.44\%\), while Urban2Vec records increases of \(4.28\%\) and \(0.90\%\). By contrast, the effects on SPPE and Space2Vec are less consistent across the two cities.

These results indicate that explicit distance is important when the input representation itself contains limited spatial relational information. In contrast, LE4Mob can support accurate flow generation without requiring distance as an additional handcrafted input. The improvement after removing distance further suggests that, once spatial separation has been incorporated into the embedding space, supplying it again as a separate feature may be redundant and could introduce competing signals into the downstream model. Nevertheless, because some other learned embeddings also remain robust without distance, this ablation should be interpreted as evidence that LE4Mob successfully encodes usable distance-related information, rather than as evidence that such information is exclusive to LE4Mob.

\section{Discussion}

\subsection{Role of Distance-Aware Representation Learning}

The results show that embedding spatial structure into the location representation can benefit mobility modelling tasks. Although many mobility models already use distance explicitly, especially in flow generation, supplying distance as a downstream OD-pair feature is different from learning embeddings whose geometry already reflects spatial relationships. The former is decoder-specific; the latter makes spatial information reusable across models.

In next location prediction, spatially structured embeddings may provide the Transformer-based model with a more informative input space for learning distance-related mobility regularities, as reflected in LE4Mob’s consistently lower spatial errors. This property is also practically relevant to location-based services and POI recommendation, where users’ destination choices are strongly distance-sensitive; when the exact destination is not identified, a geographically closer prediction can still provide a relevant and actionable alternative. In flow generation, the effect is more direct: the Bilinear model estimates OD compatibility from interactions between origin and destination embeddings, and therefore benefits strongly from embeddings that preserve spatial relationships.

The DeepGravity results further support this interpretation. When explicit OD distance is included, LE4Mob is competitive but does not always outperform all baselines, suggesting that the decoder can already exploit the supplied distance feature. When distance is removed, however, baseline representations degrade, while LE4Mob improves and achieves the highest CPC in both London and Greater Manchester. This indicates that LE4Mob preserves useful spatial information in the embeddings. Its contribution is therefore not to replace explicit distance features, but to make location embeddings more spatially self-contained and less dependent on decoder-specific distance engineering.

\subsection{Implications for Mobility Modelling}

LE4Mob separates the representation of places from the modelling of movement behaviour. Instead of learning embeddings from trajectories or OD flows, it learns from geographic context and transfers the resulting representations to downstream mobility tasks. The encoder captures where places are, what functions they serve, and how they relate spatially, while downstream models such as MHSA, DeepGravity, and Bilinear learn task-specific behavioural patterns.

This separation is particularly relevant because next location prediction and commuter flow generation differ in scale, data structure, and modelling objective. The results 
provide evidence that geography-derived embeddings can serve as a shared representational layer across individual- and population-level mobility models. This modular design also points towards future mobility foundation models in which location representation and behavioural modelling are handled by separate but connected components.

LE4Mob is also grounded in established mobility theory and recent advances in urban representation learning. Classical gravity models explain spatial interaction through the masses (population) of origins and destinations and the distance between them \cite{zipf1946p}, while DeepGravity demonstrates that incorporating richer geographic attributes through deep neural networks can substantially improve flow estimation \cite{simini2021deep}. LE4Mob follows this broader principle by encoding place functions through POI-derived semantic context and spatial relationships through distance-aware regularisation. Previous research has further shown that POI-derived urban representations can capture information associated with population and socioeconomic characteristics \cite{liu2026cityrep}. The combination of spatial semantics and distance awareness therefore provides a theoretically motivated representation of the factors shaping mobility, which may partly explain its effectiveness across the evaluated tasks.

\subsection{Limitations and Future Work}

This study focuses on next location prediction and flow generation because both directly test location representation quality. It is not intended as a benchmark across all mobility tasks. Trajectory generation and crowd flow prediction involve additional challenges, such as sequential path generation and time-series forecasting, and would require different experimental designs.

LE4Mob’s inductiveness is also evaluated within the geographic domain covered by pre-training. It can encode unseen locations in the study area, but this should not be conflated with zero-shot transfer to entirely new cities. Cross-city transfer would involve substantial domain shift in POI distributions, urban morphology, and semantic–spatial relationships. Future work could explore multi-city or national-scale pre-training for broader geographic generalisation.

Finally, LE4Mob relies on POI-derived contextual descriptions and uses a soft distance-aware regulariser. POIs are unevenly distributed, so representations may be better constrained in POI-dense areas than in sparse areas. Meanwhile, the distance objective encourages spatial consistency but does not enforce exact global distance preservation. This is a deliberate trade-off, since rigid distance preservation could weaken semantic organisation. Future work could explore spatially balanced sampling, auxiliary spatial anchors, and alternative distance formulations (e.g. travel time) and distance-aware objectives.





\section{Conclusion}

This study proposed LE4Mob, an inductive, distance-aware, and geography-derived location embedding framework for human mobility modelling. Unlike mobility-derived embeddings learned from task-specific trajectories or OD flows, LE4Mob pre-trains a location encoder from coordinates and POI-derived semantic context, while explicitly regularising the embedding space using geographic distance. By capturing both the functional characteristics of places and their spatial relationships, LE4Mob provides a reusable representation layer for downstream mobility models.

We evaluated LE4Mob on individual-level next location prediction and population-level commuter flow generation. The results show that geography-derived embeddings can transfer across substantially different mobility tasks and that representation-level distance awareness improves spatial informativeness, especially when downstream models rely directly on embedding interactions or lack explicit distance inputs. Future work will explore broader multi-city pre-training and alternative distance objectives to improve geographic generalisation. Overall, LE4Mob offers a promising step towards shared geographic representations for human mobility modelling.

\bibliographystyle{IEEEtran}
\bibliography{xlw_refs}

\vfill

\end{document}